\documentclass[sigconf, screen]{acmart}
\AtBeginDocument{%
  }

\copyrightyear{2026}
\acmYear{2026}
\setcopyright{cc}
\setcctype{by}
\acmConference[HRI '26]{Proceedings of the 21st ACM/IEEE International Conference on Human-Robot Interaction}{March 16--19, 2026}{Edinburgh, Scotland, UK}
\acmBooktitle{Proceedings of the 21st ACM/IEEE International Conference on Human-Robot Interaction (HRI '26), March 16--19, 2026, Edinburgh, Scotland, UK}
\acmDOI{10.1145/3757279.3785644}
\acmISBN{979-8-4007-2128-1/2026/03}

\usepackage{multirow}

\def\eg{\emph{e.g., }}

\newcommand{\pquotes}[1]{\textcolor[gray]{0.35}{\textit{#1}}}

\begin{document}

\title{\textit{Lantern}: A Minimalist Robotic Object Platform}

\author{Victor Nikhil Antony}
\authornotemark[1]
\affiliation{%
  \institution{Johns Hopkins University}
  \city{Baltimore}
  \state{Maryland}
  \country{USA}
}

\author{Zhili Gong}
\authornotemark[1]
\affiliation{%
  \institution{Rice University}
  \city{Houston}
  \state{Texas}
  \country{USA}
}

\author{Guanchen Li}
\affiliation{%
  \institution{Johns Hopkins University}
  \city{Baltimore}
  \state{Maryland}
  \country{USA}
}

\author{Clara Jeon}
\affiliation{%
  \institution{Johns Hopkins University}
  \city{Baltimore}
  \state{Maryland}
  \country{USA}
}

\author{Chien-Ming Huang}
\affiliation{%
  \institution{Johns Hopkins University}
  \city{Baltimore}
  \state{Maryland}
  \country{USA}
}
\thanks{*equal contribution (VNA, ZG). \textbf{Author CRediT}: Conceptualization \& Visualization (VNA,ZG,CH); Investigation \& Data Curation (VNA,ZG,GL,CJ); Method. (VNA, CH); Funding \& Supervision: (CH); Original Draft: VNA, ZG; Review \& Editing: all. \textbf{AI Use}: Text edited with LLM; output checked for correctness by authors}
\renewcommand{\shortauthors}{Antony et al.}

\begin{abstract}
Robotic objects are simple actuated systems that subtly blend into human environments. We design and introduce \emph{Lantern}, a minimalist robotic object platform to enable building simple robotic artifacts. We conducted in-depth design and engineering iterations of Lantern’s mechatronic architecture to meet specific design goals while maintaining a low build cost ($\sim$\$40 USD). As an extendable, open-source platform, Lantern aims to enable exploration of a range of HRI scenarios by leveraging human tendency to assign social meaning to simple forms.
To evaluate Lantern’s potential for HRI, we conducted a series of explorations: 1) a co-design workshop, 2) a sensory room case study, 3) distribution to external HRI labs, 4) integration into a graduate-level HRI course, and 5) public exhibitions with older adults and children. Our findings show that Lantern effectively evokes engagement, can support versatile applications ranging from emotion regulation to focused work, and serves as a viable platform for lowering barriers to HRI as a field.

\end{abstract}





\maketitle

\section{Introduction}


Robotic objects are low degree-of-freedom, actuated systems \cite{bianchini2016towards} often designed to resemble everyday artifacts such as lamps, balls, or trashcans \cite{fischer2015initiating, christiansen2018breathing}, though they may also adopt more abstract forms. Their system-level simplicity enhances long-term reliability and fault tolerance \cite{reik2017healthcare}, while their flexible design allows for easier integration into human-centric spaces, as their forms can be adapted to blend in subtly with existing environments. Moreover, people's tendency to assign meaning and social agency to objects---regardless of form \cite{duffy2003anthropomorphism}, interpreting even basic behaviors as social signals \cite{levillain2017behavioral}---enables these robotic artifacts to exhibit social characteristics beyond their mechanical limitations \cite{erel2022enhancing, ju2009approachability, sirkin2015mechanical}.

\begin{figure}[h]
\centering
\includegraphics[width=\columnwidth, alt={The figure has six blocks arranged in two rows. [first-row-first-block] features an AI generated image of asian paper lanterns that inspired the design of Lantern. The other two blocks in the first row feature the base design of lantern shown without an embodiment in deflated and inflated states. The second row shows variants of Lantern with the first block shown a robotic speaker with a blue-white-pink colored covering and light emulating from inside. The second block shows a robotic toy in a yellow covering and two googly eyes and the last block shows a robotic lamp with paper embodiment putting off red light}]{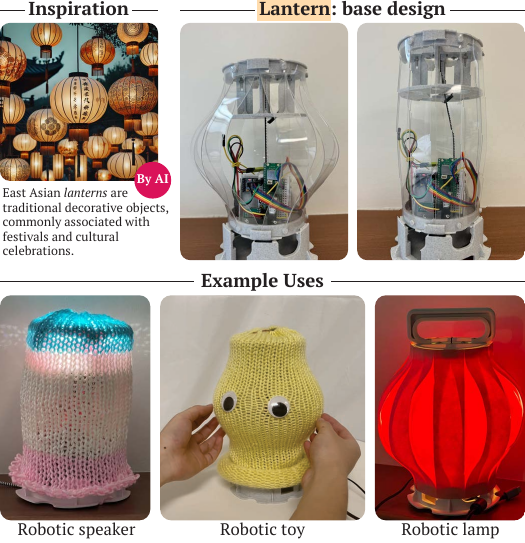}
\caption{\textbf{\textit{Lantern}} is an affordable, adaptable robotic platform inspired by East Asian paper lanterns, designed to enable the creation of simple robotic objects for diverse HRI use cases.}
\label{fig:teaser}
\end{figure}

Simple robotic objects can serve diverse and complex functions, such as inspiring creativity \cite{alves2017yolo}, facilitating human interactions \cite{kumaran2023cross, tennent2019micbot}, and supporting therapeutic interventions \cite{sabinson2024every}. These robotic objects, despite their minimal designs, can evoke perceptions of safety \cite{chakravarthi2024social}, emotional support \cite{erel2022enhancing}, and interest from users \cite{manor2022non}. This underscores their potential not only as \textit{practical} tools but also as \textit{social} actors, effectively bridging the gap between functionality and social presence. However, most existing robotic object prototypes are fully integrated, closed-source systems designed for specific applications, leaving little room for adaptation, limiting exploration of the functional and interactive potential of robotic objects. 

We introduce \textbf{\emph{Lantern}}, a minimalist robotic object platform designed as an adaptable, low-cost solution to enable a variety of human-robot interaction (HRI) explorations with robotic objects (see Fig. \ref{fig:teaser}). For instance, Lantern enables the construction of simple robotic objects that, with intentional behavior design, can act as robotic pets (see Fig. \ref{fig:lantern-extension}-b) or social robots \cite{duffy2003anthropomorphism}. This could reduce entry barriers to social HRI research, which has largely been limited to a few well-funded labs in economically developed regions. Furthermore, Lantern's low-cost design has the potential to enable larger-scale, longer-term studies of human-robot interactions. 

Lantern is engineered to be a \textit{simple}, \textit{low-cost} (unit cost is $\sim$\$40 USD) platform that is easy to assemble and adapt for different use cases (see Fig. \ref{fig:teaser} for examples). Lantern is designed to support rich physical interactions with users; its portable, \textit{holdable} form factor integrates \textit{haptic} feedback, enabling tactile engagement. To explore Lantern's interactive potential, we conducted a series of demonstrations and observational studies \cite{ledo2018evaluation} along with a workshop to imagine use cases and design interactions (Fig. \ref{fig:demonstrations}); We shared Lantern with four HRI groups and incorporated it into a graduate-level HRI course to evaluate its replicability and to gain insights into  its broader potential as an HRI platform.


This work makes the following contributions: 
\begin{itemize}
    \item We design, develop and open-source our system: Lantern, a minimalist, haptic robotic object platform.
    \item We explore Lantern in various interaction settings to understand its potential for research and broader HRI.
    \item We provide insights from our explorations of Lantern to inform the future design of robotic objects.
\end{itemize}

\section{Related Works}

\subsection{Robotic Objects and Their Uses}
A broad array of simple robotic objects have been designed to execute functions well beyond their minimalist form \cite{levillain2017behavioral}. For example, \textit{Micbot}, a simple actuated microphone, influenced conversational dynamics in group settings. By employing non-verbal cues to mimic back-channeling behaviors, it enhanced team performance and engagement \cite{tennent2019micbot}. Even robotic furniture (\eg mechanical ottoman) can communicate intentions through movement alone, demonstrating how simple motion can convey purpose \cite{sirkin2015mechanical, nguyen2025eliciting}.

Abstract robotic objects with low degrees of freedom (DOF) actuation have been successfully used to convey social intentions, such as greetings. These robots have also positively impacted social perceptions, particularly trust and safety in human-robot interactions \cite{anderson2018greeting, chakravarthi2024social}. Extending beyond social signaling, robots like YOLO and Ranger leverage simple autonomous behaviors to foster creativity in children and promote practical behaviors like tidying up during playful interactions \cite{alves2017yolo, fink2014robot}. This highlights how minimalistic robotic designs can inspire meaningful human engagement.

Incorporating haptic feedback in robotic objects introduces multi-sensory interactions enriching human-robot communication. \textit{Haptic Creature} and \textit{EmotiBot} exemplify this by allowing for affective, touch-based interactions, which foster more natural and emotionally resonant connections between humans and robots \cite{thompson2024emotibot, yohanan2012haptic}. In addition to promoting emotional engagement, physical interactions with robotic objects have been employed to address complex needs, such as pain communication and behavioral therapy \cite{kim2023alhe, sabinson2024every}.

\textbf{\textit{Lantern}} aims to offer an open-source platform that enables exploration of the creative potential of simple robotic objects while supporting rich, multi-sensory haptic interactions.

\subsection{Open-Source Robots}
Open-source platforms have broadened access to HRI research, but differ significantly across interaction modalities, physical form, and accessibility. Robots like Blossom\cite{suguitan2019blossom}, Ono\cite{vandevelde2013systems}, and FLEXI\cite{alves2022flexi} support expressive motion and appearance customization but lack haptic feedback and are not designed for handheld use. Humanoid platforms such as Poppy\cite{lapeyre2014poppy}, Reachy\cite{mick2019reachy} and iCub\cite{metta2010icub} offer richer  interaction and sensing but are expensive, stationary, and complex to build or modify. Ommie \cite{matheus2025ommie}, a compact, affect-oriented robot, supports vibro-tactile modality, yet remains closed-source limiting its accessibility to researchers and HRI practioners.



\textbf{\textit{Lantern}} addresses these limitations by offering an affordable, holdable, open-source robotic platform that integrates vibro-tactile and kinesthetic haptic feedback, enabling researchers to explore rich, tactile human-robot interactions at an affordable cost.

\begin{table}[h]
\centering
\caption{Comparison of Robotic Platforms}
\setlength{\tabcolsep}{3pt} 
\begin{tabular}{@{}lp{1.55cm}cccc@{}}
\toprule
\textbf{Robot} & \textbf{Type} & 
\rotatebox{45}{\textbf{Cost(\$)}} & 
\rotatebox{45}{\textbf{Haptic}} & 
\rotatebox{45}{\textbf{Holdable}} & 
\rotatebox{45}{\textbf{Custom.}} \\ 
\midrule
Yolo \cite{alves2017yolo} & Robotic Obj. & 150-200 & - & - & - \\
Blossom \cite{suguitan2019blossom} & Social Robot & $\sim$250 & - & - & $\bullet$ \\
ERGO \cite{desprez2018poppy} & Robot Arm & 300 & - & - & - \\
Ono \cite{vandevelde2013systems} & Social Robot & 350-575 & - & $\circ$ & $\bullet$ \\
Flexi \cite{alves2022flexi} & Social Robot & $\sim$2500 & - & - & $\bullet$ \\
Quori \cite{specian2021quori} & Mobile Robot & $\sim$6300 & - & - & - \\
\midrule
\textit{Lantern} & Robotic Obj. & $\sim$40 & $\bullet$ & $\bullet$ & $\bullet$ \\
\bottomrule
\end{tabular}
\smallskip 
\footnotesize
\textbf{Legend:} $\bullet$ = Full feature, $\circ$ = Partial feature, - = No feature
\label{tab:robot_comparison}
\end{table}



\section{Design Goals}
\textbf{\textit{Lantern}} is engineered with an \textit{``easy assembly, warm interaction''} mindset at its core, employing minimalist design principles. Prioritizing affordability, adaptability, and ease of assembly, Lantern is a step towards lowering barriers to HRI research with a low-cost, easy-to-build robotic object platform\footnote{Lantern's build guide, and code (Section \ref{sec:software}) can be found at our project site: \url{https://lantern-website-iota.vercel.app/}}. The following design goals guided Lantern’s development to achieve this vision:
 
\textbf{Haptic.} Haptic interaction forms the basis of Lantern's ability to create engaging interactions through low degree-of-freedom actuation. Through incorporating vibrotactile and kinesthetic haptic feedback mechanisms, Lantern enhances the tactile experiences that can be crafted, while opening up new possibilities for affect communication across various interaction contexts \cite{yohanan2012role}.

\textbf{Holdable.} To expand the range of human-robot interaction possibilities, we design Lantern as a holdable and portable platform. Beyond simply its compact and light form factor, its soft exterior and deformable shell embolden its holdable profile. Moreover, Lantern can communicate via Wi-Fi and be powered with a battery, further enhancing its portability and ability to support diverse interactions.

\textbf{Low-Cost.} With a target per-unit build cost of around \$40 USD, Lantern aims to lower the economic barriers to entry for interactive robotics research, particularly to enable in-the-wild, large-scale, long-term human-robot interaction studies.

\textbf{Simple.} A minimalist design for Lantern aims to streamline its mechanics and control requirements. This simplicity aims to make the robot more reliable, accessible, and less intimidating to build and adapt by reducing system complexity when possible. Lantern features simplified fabrication and assembly processes to facilitate rapid replication and further lower barriers to entry towards making Lantern an accessible platform for education and research contexts.

\begin{figure}[t]
\centering
\includegraphics[width=\columnwidth, alt={Illustration of the design iteration of the Lantern - three images in a single row. The first image shows a 3D printed prototype  six guiding rods and a servo-driven belt system  and a middle chassis. The second image shows an updated verison of the prototype with an integrated base. The last image shows the final version of lantern with the anti-collapse brace and snap-on stand highlighted}]{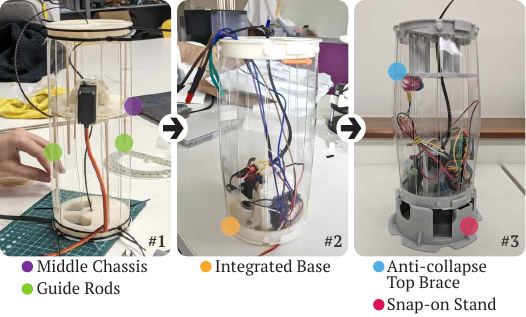}
\caption{Lantern underwent several design iterations. \textit{\#1}: Initial design had six guide rods and a servo-driven belt system but suffered from tilting and incomplete strokes due to misalignment. \textit{\#2}: A major iteration replaced the guide rods with a belt guide and introduced integrated base housing. \textit{\#3}: Final design added structural braces and a stand for stability.}
\label{fig:design-iterations}
\end{figure}

\section{The Design of Lantern}
The design of \textbf{\textit{Lantern}} is inspired by traditional East Asian paper lanterns. Its core movement involves smooth, rhythmic expansion and contraction, evoking the calming motion of a floating lantern while providing haptic feedback (Fig. \ref{fig:teaser}). To achieve this, Lantern underwent multiple design iterations (see Fig. \ref{fig:design-iterations}), addressing both mechanical and electronic challenges to meet our design goals. Details on the design iterations can be found in the Appendix.


\subsection{Mechanical Design}
\begin{figure}[b!]
\centering
\includegraphics[width=\columnwidth, alt={Exploded view of Lantern’s CAD model, illustrating the key  components arranged in a column. The first group of elements feature The PET-G shield forms the exterior, supported by a top brace to prevent inward collapse. The second layer shows base housing containing the e-component tray and servo motor that drives the GT2 belt to produce the expansion and contraction movement; it also shows the pulley stand with a belt guide ensures perpendicular force transmission. The last layer features the snap-on stands provides stable base and optional battery storage.}]{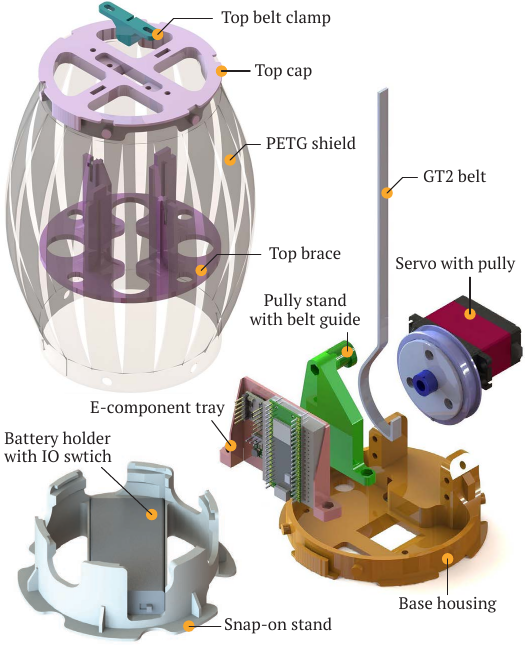}
\caption{Exploded view of Lantern’s CAD model, highlighting the base housing with servo-driven GT2 belt, the PET-G shield exterior with top brace, the pulley and belt guide for alignment, and the snap-on stand for stability}
\label{fig:mechanical-design}
\end{figure}

Lantern features a base housing connected to the top cap via a compliant plastic shell, formed by attaching a sliced PET-G sheet around the cap and housing (see Fig. \ref{fig:mechanical-design}). The base housing neatly organizes the electrical components and actuation mechanism. A servo motor, rigidly attached to the base housing and equipped with a pulley, actuates Lantern by winding and releasing a belt that links the base housing and top plate. The belt converts the pulley’s torque into a linear force, adjusting the distance between anchor points and altering the axial length; this deforms the shield and alters the Lantern’s external dimensions. A belt guide is integrated with the pulley bracket, redirecting the pulley direction to ensure perpendicular pull force is transmitted to the center of the top cap. To further ensure the belt remains perpendicular and prevent cap from tilting during actuation, the top belt clamp is adjustable, allowing it to slide into the optimal position. The top plate integrates a brace that prevents the PET-G shell from collapsing inwards when excessive exterior pressure is applied. As an accessory, a snap-on stand has been designed allowing Lantern to stand stably and have the battery pack stored internally, if needed.


\textbf{Shell.} Slicing the PET-G sheet in intervals reduces the shell's compressive strength, allowing radial expansion when the shell is axially compressed. A 1\,mm thick PET-G sheet, with 18 slices, was selected for the shield based on iterative testing, demonstrating the ability to withstand inward compression during user interaction while remaining deformable with the torque of the servo motor. 

\textbf{Modifications.} Adjustments to the pulley belt and shell length allow for easy modification of Lantern's height and expansion width, enabling the system to be tailored to different contexts. Shells of different lengths exhibit distinct responses to compression. A detailed kinematic analysis is provided in the Appendix. Additionally, varying the number of interval slices in the lantern's plastic shell enables further customization of the tactile feedback. Moreover, custom attachments (\eg carry handle for robotic lamp, see Fig. \ref{fig:teaser}) can be 3D printed and fastened to the top cap.

\textbf{Fabrication.} The entire skeletal structure of Lantern, except for the shield, can be 3D printed with PLA filament and assembled using M3 fasteners. Each build requires around 150\,g of filament. The 1\,mm PET-G sheet can be laser-cut, achievable using a hobby-grade laser-cutting machine equipped with proper exhaust features.

\subsection{Form Factor and Embodiment}

Lantern's embodiment features an exterior covering that envelops its internal skeleton. This covering can be crafted from various materials (\eg paper, yarn, cotton) with different textures and colors, allowing Lantern to be customized to suit specific applications or personalized to individual users. The use of diverse materials not only alters Lantern’s visual appearance but may even enhance the tactile experience, adding another sensory dimension for the user.
 
\subsection{Electrical Components}
Lantern is powered with  off-the-shelf electrical components (see Appendix C), aligning with our aim of creating a low-cost, easy-to-build bot by reducing material costs and part sourcing difficulty.

\textbf{Computation.} A \textit{Raspberry Pi Pico W} is Lantern's computational unit that controls its behaviors, synchronizing Lantern's actuators and sensors while facilitating communication. The Pico W offers various connection options, both wired and wireless, and ample GPIO pins several of which are intentionally unoccupied, allowing for future development and straightforward modifications.

\textbf{Actuators.} Lantern is equipped with two sets of onboard actuators to generate kinesthetic and tactile interactions. A 25 kg-cm hobby-grade servo drives the pulley and belt system, compressing Lantern longitudinally to create a radial expansion, which produces the kinesthetic expansion effect. A mini-vibration motor is attached to the system to produce the vibro-tactile feedback. A 2-channel motor driver powers the vibration motor, providing sufficient current for fast response and robust output, enabling rich haptic patterns.

\textbf{Power.} Lantern operates on any 7.5\,V DC power source with an output of 10\,W or higher and a standard 5.5\,*\,2.5\,mm plug, achievable by either a 2\,S Lithium battery pack or an AC to DC power adapter. The voltage regulator and circuit design then step down the input voltage to 5.5\,V for low-voltage components (\eg Pico) while simultaneously supplying the required input voltage directly to high-voltage components (\eg servo).  Additionally, with the 3.3\,V power output from the Pico W, Lantern’s three internal power supply options make it compatible with a wide range of add-ons. 


\subsection{Software Architecture}
\label{sec:software}
We engineered Lantern's control software to orchestrate its behaviors including its kinesthetic and vibro-tactile behaviors. We provide two different programming methods for Lantern: a microPython Software Development Kit (SDK) meant for quick prototyping and stand-alone programs and a microROS+ROS2-based system developed for enabling ROS-based interaction programming.

\subsubsection{Lantern SDK} We developed \textit{Lantern SDK} as a modular framework that provides a unified API for controlling Lantern's actuation, sensing and behaviors. Designed with extensibility in mind, the SDK allows developers to define custom behaviors, schedule actions, and respond to sensor events with minimal boilerplate. It supports both local execution and remote control over Wi-Fi, enabling quick development of expressive, event-driven interactions.

\subsubsection{microROS system} ROS is widely adopted in HRI research as the standard software stack. To ensure Lantern can integrate seamlessly into these workflows, we developed a microROS-based control system. This setup enables the Pico W to connect directly with ROS2 while offloading heavier computation. Our implementation has two layers: a microROS program with nodes executing actions on Lantern, and a ROS2 program that orchestrates these behaviors in response to higher-level events (\eg inputs from ReactJS). We demonstrate this system in section \S\ref{sec:sensory-room}.

\section{Customizing Lantern: Imaging Use Cases} 

To demonstrate Lantern’s adaptability and its viability across contexts, we integrated sound, light outputs and, IMU, and touch sensing (Fig.~\ref{fig:lantern-extension}) to enable the following use cases:

\begin{figure}[h]
\centering
\includegraphics[width=\columnwidth, alt={Illustration of the design varations of the Lantern - two storyboards illustrate two variants each in its own row. The row has a series of 4 images showing a lantern in a dark room with white covering an IMU sensor, the second frame show yellow light slowly eminating, the third shows light getting brighter and lantern inflating and last shows a pair of hands flipping a glowing lantern to turn it off. The second story board image shows Lantern in a fluffy, soft yellow covering with 3 printed ears attached to it copper capacitative strips attached to its exterior support touch sensing. The middle two frames show a hand touching Lantern in different spots and the last frames show how lantern moves when touched via overlayed motion lines.}]{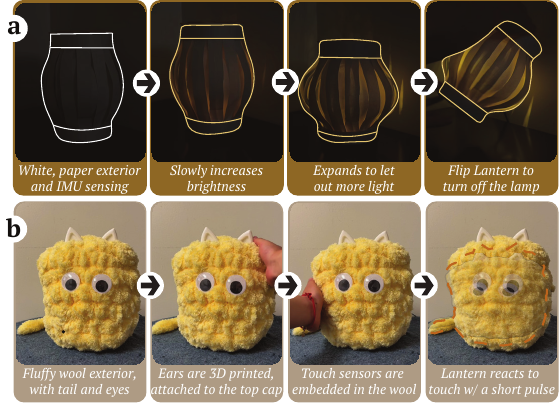}
\caption{Illustrations of Lantern’s use cases. (a) As a Circadian rhythm lamp, it guides wake-up routines. (b) As a fluffy toy with touch sensing, it can be a cuddly companion.}
\Description{Illustration of the design varations of the Lantern - two storyboards illustrate two variants each in its own row. The row has a series of 4 images showing a lantern in a dark room with white covering an IMU sensor, the second frame show yellow light slowly eminating, the third shows light getting brighter and lantern inflating and last shows a pair of hands flipping a glowing lantern to turn it off. The second story board image shows Lantern in a fluffy, soft yellow covering with 3 printed ears attached to it copper capacitative strips attached to its exterior support touch sensing. The middle two frames show a hand touching Lantern in different spots and the last frames show how lantern moves when touched via overlayed motion lines.}
\label{fig:lantern-extension}
\end{figure}

\textbf{Robotic Lamp for Circadian Well-Being:} Lantern can be adapted to be a bedside wake-up light by adding a 60-pixel RGB LED strip (\$8) and a 6-DoF IMU (\$4) (see Fig. \ref{fig:lantern-extension}.a). Thirty minutes before the wake-up alarm, the LEDs begin a gradual dawn ramp—from deep red to daylight yellow, while the breathing tempo gently accelerates. At the alarm time, the shell reaches full brightness and the vibration motor issues a soft heartbeat every 5s. The alarm is dismissed only after the user picks up Lantern and performs two gentle tilts or a single 180° flip, gestures recognized by the IMU; this light physical interaction could help counter sleep inertia while exploiting Lantern’s holdable form factor. 

\textbf{Embodied Speaker:} We extend Lantern into a mobile, vibrotactile speaker by integrating a mini speaker with microSD storage (\$6). Lantern plays locally stored audio tracks while synchronizing its expansion-contraction cycles to the beat of the music. A simple beat-detection algorithm modulates the servo’s speed and range, creating rhythmic inflation patterns that mirror the music’s tempo. Simultaneously, the vibration motor pulses in time with bass-heavy beats, adding a haptic dimension to the auditory experience. This configuration transforms Lantern into a soft, kinetic speaker that augments passive listening with active, touchable motion, offering a richer, multi-sensory musical experience.

\textbf{Soft Toy:} Lantern can be a cuddly comfort object by slipping a fluffy wool exterior over the shell and attaching two 3-D–printed TPU “ears” (\$0.5) to the top-cap (see Fig. \ref{fig:lantern-extension}.b). The servo continues its slow breathing cycle, but the on-board vibration motor now delivers low-amplitude, 20 Hz ``purr'' bursts that fade in and out every 10 s imitating the calming oscillations common in therapeutic plush toys. The ears are elastically hinged to the top cap, each expansion makes them wiggle slightly, reinforcing liveliness without adding actuators. This cosmetic add-on with the stock vibrotactile channel turns the base robot into a soft, holdable companion.

\section{Exploring Lantern's Potential}
To probe further use cases Lantern can support, we conducted a design workshop, followed by a case study examining one emerging application: \textit{emotion regulation}. We shared the platform with four external HRI groups to gather perspectives on its potential, which inspired its integration into a \textit{Introduction to HRI} course as an educational tool; We held public exhibitions to gain further insights into opportunities with special populations highlighted by external researchers. This section details our exploratory work with Lantern.

\subsection{Design Workshop: Hacking Lantern}\label{sec:hackathon}
We hosted a hackathon-style design workshop with five participants from different academic fields (\eg Mechanical Engineering, Mental Health Counseling, Human-Robot Interaction). The workshop aimed at engaging domain experts in \textit{brainstorming} potential use cases for Lantern, and included scenario-based \textit{design} and \textit{implementation} of imagined interactions. The workshop lasted two hours, and participants were compensated with a \$20 gift card as approved by our institutional review board.

\textbf{Workshop Protocol.} 
After obtaining informed consent, the session began with a hands-on demonstration of Lantern to introduce participants to its capabilities, form factor, and tactile nature. We showcased slow, controlled inflation-deflation cycles and demonstrated the vibration feedback. Following the demonstration, participants took part in a collaborative brainstorming exercise, sharing their initial impressions and discussing potential application areas for Lantern. This brainstorming stage laid the foundation for the subsequent design exploration.

Participants then self-organized into two teams, each selecting an application area identified during the brainstorming session. Using structured worksheets \footnote{link to worksheets: \url{https://tinyurl.com/2z4sz6zs}}, the teams designed interaction scenarios, discussed the context of use, and designed key behaviors for their chosen applications. They story-boarded the interaction flow and conceptualized the ideal embodiment of Lantern for their contexts. Finally, each team programmed and demonstrated the core behaviors for their interactions, building on the base program used during the initial demonstration.



\subsubsection{Imagined Applications} 
Participants identified several potential applications for Lantern, including its use in breathing exercises for training or recovery, assisting with emotion regulation (\eg anxiety relief), supporting yoga or Pilates routines, providing encouragement during spirometry tests, and aiding in maternity delivery training. They also envisioned Lantern as a coping mechanism for managing psychological stressors or disorders. This highlights how simple robots built with the Lantern base could enable a wide range of applications. Two teams designed interactions for distinct use cases: \textit{Emotion Regulation for Children in Schools} and \textit{Post-Operative Breathing Exercises in Hospitals}. Notably, both teams successfully implemented key behaviors of envisioned interactions on Lantern in less than 20 minutes, showing the ease of adapting our system.

\subsubsection{Emotion Regulation for Children in Schools} Lantern was envisioned for use in elementary school sensory rooms, particularly after conflicts, to help children self-regulate their emotions through guided breathing exercises. Breathing behaviors were designed for Lantern to guide the child’s breath, gradually slowing it down until the child could self-regulate. When the child is able to self-regulate, the breathing behavior would stop, and the heartbeat behavior would begin; the behaviors are triggered one at a time, avoiding overlap to prevent sensory overload. Instructions on how to interact with Lantern would be in the form of pictorial guides or videos.

\textit{Behaviors.} Two distinct ``breathing patterns'' were implemented to aid in emotion regulation:      ``\textit{Bunny breathing},'' consisting of short, quick breaths through the nose, and “\textit{Dragon breathing},” characterized by slow, deep inhales and exhales. A “steady, slow-paced” heartbeat pattern was proposed to help further calm and comfort children after the breathing exercises.

\textit{Embodiment.} Participants suggested a soft, cuddly material in bright colors, resembling stuffed animals, for Lantern’s embodiment. They also recommended that the design resemble cartoon characters or animals, with the addition of a fidget tray to further support emotion regulation. Additionally, the embodiment (outfit) should be easy to remove and washable.

\subsubsection{Post-Operative  Exercises in Hospitals} Lantern was envisioned for use in hospital rooms post-surgery to assist patients with guided breathing exercises. Each patient would be assigned their own Lantern, introduced by a healthcare professional. Participants suggested that Lantern could interface with existing medical devices, such as spirometers, to track progress and adapt its breathing protocols over time. Additionally, participants imagined Lantern providing verbal feedback to motivate patients and offering a hand massage through gentle vibrations after each session. Lantern could also support breathing exercises as part of in-home care.

\textit{Behaviors.} A standard post-operative breathing exercise\cite{townsend2016sabiston} was successfully implemented on Lantern, involving a sequence of three normal breaths followed by one slow, deep inhale (3--5 seconds) and a controlled exhale. Each set of three normal breaths and one deep breath constitutes a repetition.

\textit{Embodiment.} Participants proposed a soft, silicone exterior to give Lantern a fluid, responsive feel. For pediatric settings, they envisioned a fuzzy outer coating to make Lantern more appealing.

\subsection{Case Study: Lantern in a Sensory Room} \label{sec:sensory-room}
To examine Lantern as a proof-of-concept in open-ended interactions grounded in a practical use case, we conducted a case study inspired by the \textit{Sensory Room for Emotion Regulation} concept designed in our \emph{Hacking Lantern Workshop} (Section \ref{sec:hackathon}).

\begin{figure*}[t]
\centering
\includegraphics[width=\textwidth, alt={Shows the series of interactions featuring graduate students, older adults and children that were used to explore Lantern. There are two rows. The first row has five images. The first two images shows a Lantern variant built by students of an intro to HRI course to be a study buddy for people with ADHD - it is dressed like a Owl holding a pen and notebook and glasses that can be lowered to turn it on. The second third depicts the sensory room setup with a glowing humifier and a Lantern with yellow covering. The fourth image shows a public exhibtion with older adults at a senior living center with 5 older adults and one older adult holding a grey lantern. The fifth image shows two kids interacting with lantern and one child is holding the lantern in his lap. The second row features five images showing the range of interaction in the sensory room with the first image showing a person holding Lantern in their lap, second and third image showing people touch lantern while its on a table, the fourth image shows a person holding lantern close to their face and talking to it and the last image shows a person turning latern around and poking it on the table while leaning back on their chair.}]{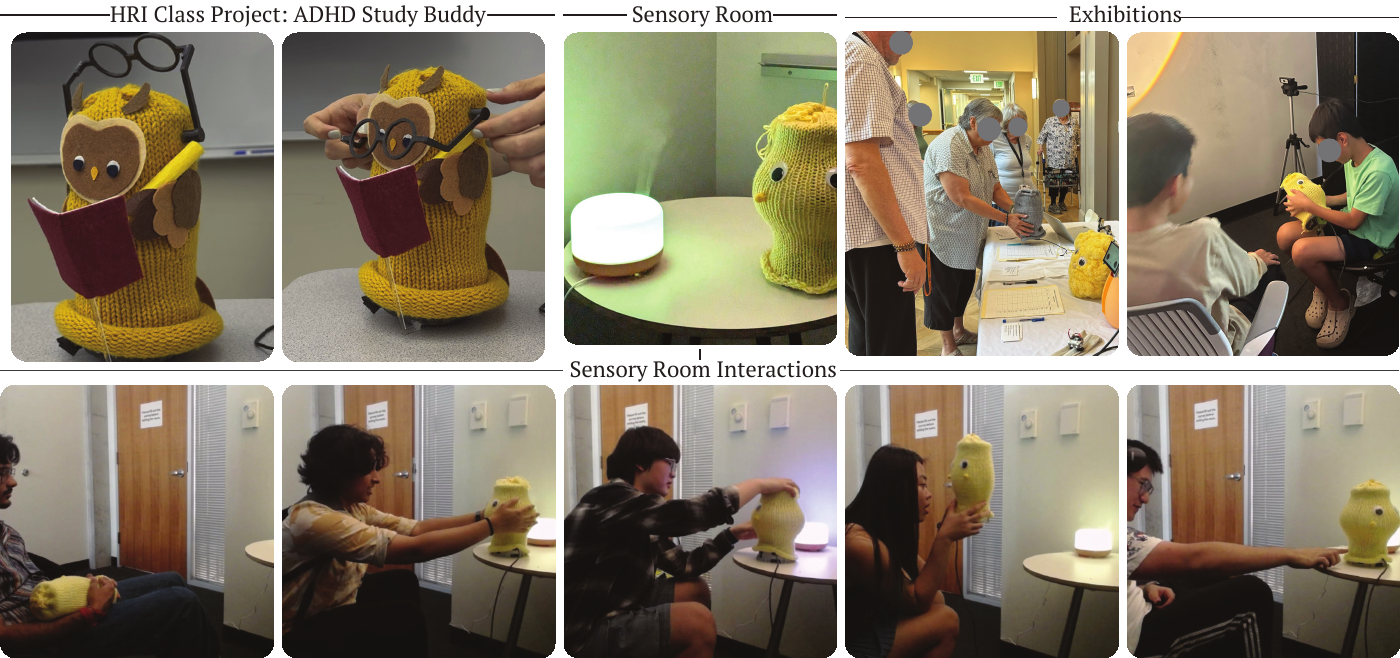}
\caption{We explore Lantern's potential through a series of interactions featuring graduate students, older adults and children.}
\Description{Shows the series of interactions featuring graduate students, older adults and children that were used to explore Lantern. There are two rows. The first row has five images. The first two images shows a Lantern variant built by students of an intro to HRI course to be a study buddy for people with ADHD - it is dressed like a Owl holding a pen and notebook and glasses that can be lowered to turn it on. The second third depicts the sensory room setup with a glowing humifier and a Lantern with yellow covering. The fourth image shows a public exhibtion with older adults at a senior living center with 5 older adults and one older adult holding a grey lantern. The fifth image shows two kids interacting with lantern and one child is holding the lantern in his lap. The second row features five images showing the range of interaction in the sensory room with the first image showing a person holding Lantern in their lap, second and third image showing people touch lantern while its on a table, the fourth image shows a person holding lantern close to their face and talking to it and the last image shows a person turning latern around and poking it on the table while leaning back on their chair.}
\label{fig:demonstrations}
\end{figure*}

\subsubsection{Case Study Setup.} A sensory room was set up in a campus building, designed to provide a soothing, multi-sensory experience. The space featured soft lighting, slow lofi background music, and a humidifier dispersing lavender essential oils to enhance the calming atmosphere. Lantern was placed on a small table beside a soft saucer chair and programmed with a slow breathing behavior (Fig. \ref{fig:demonstrations}). 

\textbf{Protocol.} Before entering the sensory room, participants provided informed consent. Instructions for starting the sensory room were posted on the door. 
After their session, participants filled out a short digital survey consisting of four 5-point Likert scale questions and an open text-response about their experience (see Appendix). All interactions with Lantern were video recorded.

\textbf{Participants.} We advertised the sensory room via flyers and community group chats to invite voluntary interactions. A total of 20 participants (13M/7F, aged 18--31) interacted with Lantern in this study approved by our institutional review board (IRB). 

\textbf{Analysis.} Two researchers independently coded the videos for interaction patterns and agreed on all codes through discussion. Two participants exited the room without completing the survey and were excluded from the analysis.

\subsubsection{Findings and Observations}
On average, participants interacted with Lantern for around 5 minutes, with individual interaction times ranging from 1.5 to 14.3 minutes. Participants generally responded positively to their interactions with Lantern, with most agreeing that the experience had a calming effect. When asked to rate the statement, “I am feeling calmer now than before I interacted with the robot,” participants gave an average score of 4 ($SD=0.59$). For example, P7 remarked, \pquotes{``Interaction with the robot felt like interacting with a pet. It was really nice and helped me relax.’’} Additionally, two participants referred to Lantern as a “companion,” reflecting its potential to evoke a sense of comfort and companionship.

Overall, the interaction with Lantern was positively received, with participants rating their experience with the robot at an average of 4.17 ($SD=0.62$). As P16 noted, \pquotes{``The [robot's squeaking] with its appearance was kind of charming, and I spent a bit of time observing its movements, which helped me focus more on my thoughts.''}

\textit{Perception of Liveliness:}
Participants’ perception of the robot’s liveliness was mixed, with an average score of 3.11 ($SD=1.08$). Some participants attributed low liveliness to factors such as slow movement speed (P4), predictability (P10), and a lack of behavior such as eye movements (P17).

\textit{Interaction behaviors:} Participants demonstrated a wide range of interaction behaviors with Lantern, reflecting diverse approaches to engaging with the robot (see Appendix E). Some gently petted or pressed Lantern’s head, while others held it in their palms, placed it on their laps, or held it while it was on the table. Interaction styles varied from soft touches and pokes to more forceful squeezes, showcasing different levels of tactile exploration (see Fig. \ref{fig:demonstrations}). In terms of engagement, some participants leaned back and observed Lantern quietly while smiling or staring at the robot as it moved. Others even engaged verbally, occasionally talking to it. The wide spectrum of observed interaction patterns illustrates how Lantern fosters passive observation and active interaction.


\subsection{Lantern as a Research Tool}
Lantern aims to support HRI research by fostering an open-source approach to robotic objects. As a step towards sharing Lantern with the wider HRI community, we provided four external HRI labs with a build guide, an instructional video, and a kit containing the materials necessary to construct Lantern. We then hosted online discussions with two of the groups to learn about their building experience and their perspective on how Lantern may be used for HRI research. We received informed consent from the researchers (n=5) and compensated them at \$15USD/hr as approved by our IRB.

The two groups successfully replicated Lantern within 2--4 hours. Participating researchers suggested a variety of potential use cases for Lantern in HRI research. For example, R1 proposed using Lantern to introduce haptic sensations to existing robotic platforms, stating that Lantern could be \pquotes{``a pair of lungs on [the QT robot]… add a degree of freedom that’s salient, but not expensive or too difficult.’’} 



R2 highlighted Lantern’s unique design as a source of interesting HRI challenges, particularly in relation to its ability to change shape: \pquotes{``[Lantern] is a different kind of robot than most of the ones I’ve seen because it changes shape, whereas most robots are pretty rigid. It would be an interesting problem to think about how to design exteriors for underlying robots, because you have to consider how things stretch.’’}

The researchers were also interested in exploring affective communication through Lantern, suggesting the addition of speakers, LEDs, or the use of movement to convey emotion. Further enhancements, such as adding sensing capabilities to detect emotions, respond to sentiment or tone, or simply sense a person’s presence, were proposed to expand Lantern’s potential applications. As R3 expressed, \pquotes{``I would like to put it just next to my door and when I go back home, it just uses its gestures to show ‘I’m happy you’re here.’’’}

Beyond research, participants also proposed using Lantern in educational settings for undergraduates and children. R4 and R5 suggested that older adults could build a system like Lantern as part of a cognitively stimulating learning process, while also fostering a personal connection through co-building. As R5 noted, \pquotes{``it can be empowering if people build it from scratch… there’s a lot of learning they can do, and the design afterwards can be a very fun activity.’’}

\subsection{Lantern as an Educational Platform}
Feedback from external researchers, combined with the lack of low-cost, adaptable platforms for HRI education, motivated us to explore Lantern’s role as a teaching tool. We integrated it into a graduate-level \textit{Introduction to HRI} course with 30 students organized into ten teams. The course provided a broad, hands-on experience, from early ideation to full research projects. Each team received a Lantern kit, assembled it, and demonstrated core functionality before advancing to project work. This on-boarding process helped students gain confidence in working with Lantern's hardware and code.

Early in the semester, students engaged in co-design exercises to imagine applications for Lantern, informed by HRI papers provided as course material. The final project then asked them to explore these ideas through substantial, research-driven investigations. Teams pursued different pathways: participatory design, controlled experiments, computational methods, or system development; choosing approaches that matched their interests while engaging with the methodological spectrum of HRI.

Building on their concepts, teams experimented with behaviors and embodiments, using Lantern’s simple, tactile platform to explore how physical form and motion could convey intent, emotion, or utility. Its design encouraged rapid iteration, allowing students to move fluidly from sketches to working prototypes.

As a result, projects spanned diverse domains: participatory design sessions with peers, interaction studies probing trust and engagement, and prototypes tailored for specific contexts. One team, for example, developed a ``\textit{study buddy}'' for students with ADHD based on body doubling principles (see Fig.\ref{fig:demonstrations}), while others built therapeutic companions or examined interaction transparency. The combination of co-design, building, and research exploration highlighted Lantern’s versatility as a teaching tool, enabling students to engage with the full HRI pipeline—envisioning, prototyping, experimenting, and analyzing—while infusing their own creativity.

\subsection{Public Exhibitions and Engagement.}
To further probe what kind of interactions Lantern evokes with special population (\eg children, older adults), as discussed by external researchers, we conducted a series of public exhibitions to inform future applications and interaction design.

\textbf{Older Adults Community Center.}
We exhibited Lantern at an outreach session in a senior living center. Older adults were invited to engage with various interactive systems, including Lantern. During the one-hour session, Lantern was programmed to perform slow inflation-deflation cycles, mimicking a calming, slow-breathing pattern.
Participants interacted with Lantern in a variety of ways: \textit{holding, lifting}, and even \textit{squeezing} it (Fig. \ref{fig:demonstrations}). Overall, the robot was met with positive feedback, with many participants describing Lantern as ``\textit{adorable},'' ``\textit{cute},'' and ``\textit{neat}.'' A notable observation during the exhibition was a recurring tendency for older adults to squeeze Lantern tightly. Despite not being explicitly designed for squeezing, Lantern consistently  recovered from these interactions, showcasing its robustness and resilience under a variety of human handling. This behavior highlighted not only the system’s durability but also the potential for designing future iterations that specifically support squeezable, tactile interactions in Lantern.

\textbf{Children's Robotics Open-House.}
We hosted a robotics open-house for children aged 6--12 years (n=8) to experience interactive robots. At this event, we demonstrated the ``sensory room'' setup, where children were invited to interact freely with Lantern, which was running a slow-breathing behavior. 

The children exhibited a range of interaction patterns. Some handled Lantern gently, treating it with noticeable care, while others engaged more playfully, poking the robot or holding it upside down. Some children remarked that they felt relaxed or even sleepy after interacting with Lantern, with one child suggesting that the robot could help calm down peers experiencing anger issues at school, reflecting a concept explored in the design workshop.

There was significant curiosity about how the robot functioned, prompting us to remove the outer cover and provide explanations of its components and operation. These observations suggest that Lantern’s design not only encourages varied forms of engagement but also sparks curiosity and reflection among younger users, indicating potential on using Lantern to promote STEM education.

\subsection{Key Takeaways}

\textbf{\textit{Lantern}} serves as a flexible, low-cost platform that supports the creation of \textit{simple}, \textit{haptic}, and \textit{holdable} robotic objects with a range of potential applications. Through introducing Lantern via a design workshop, a case study, sharing with HRI groups, an academic course, and public exhibitions, we have demonstrated that Lantern effectively meets its core design goals. These interactions not only highlight the strengths of the platform but also reveal opportunities for enhancing its interaction capabilities, paving the way for future research and broader applications of robotic objects in HRI.







\section{Design Opportunities}
Across contexts, interactions with Lantern surfaced unexpected design tensions, surfacing new design possibilities we present below.

\subsection{Enriching Interactions With Robotic Objects}

Our observations of interactions with Lantern revealed several promising avenues for enriching user engagement through sound, verbal communication, and tactile feedback.

\subsubsection{Sound Design} Participants often described the operational sounds from Lantern (e.g., the hum of its motor) as soothing and charming. This highlights the potential for intentional sound design to enhance interactions with robotic objects \cite{zhang2023nonverbal}. Incorporating rhythmic sounds, like a soft “lub-dub” heartbeat paired with tactile vibrations, can create a more immersive, multi-sensory experience \cite{robinson2022designing, jee2010sound}. Additionally, reactive sounds could make Lantern’s responses feel livelier, such as a calming tone when touched or a playful chime when shaken. Synchronized ambient sounds, such as ocean waves or rainfall, with Lantern’s movements, could further enrich interactions, particularly for therapeutic or relaxation uses \cite{iyendo2017sound}. For instance, during guided breathing exercises, the sound of waves could align with Lantern’s inflations and deflations, offering a calming, immersive experience.

\subsubsection{Language Capabilities} We observed that Lantern can invite verbal interactions. Beyond ambient sounds, integrating verbal communication capabilities offers potential for deeper engagement. By utilizing tiny language models \cite{hillier2024super, jiao2019tinybert, correa2024teenytinyllama, cohn2023eelbert}, Lantern could provide simple verbal responses grounded in the context of its role. However, introducing complex verbal interactions could raise user expectations, potentially clashing with Lantern’s minimalist form and leading to a disjointed user experience and undermine the human-robot relationship \cite{bartneck2009my, walters2008avoiding}. Introducing verbal interactions that feel natural and aligned with the simplicity of robotic objects, without creating unrealistic expectations or diminishing the subtlety of the design is an open challenge.

\subsubsection{Tactile Engagements} Our findings also suggest a strong desire for more tactile interactions. Although Lantern was not initially designed for squeezing or compressing, many participants, particularly older adults, instinctively tried to engage with it in this way. This behavior indicates that Lantern’s form naturally invites such interactions. To support a broader range of tactile experiences, we propose incorporating flexible, 3D-printed shells made of TPU (Thermoplastic Polyurethane), an elastic and compressible material. This would allow users to squeeze Lantern, enabling a richer haptic experience and meeting user expectations. Such enhancements would expand Lantern’s applicability in therapeutic and sensory contexts \cite{krichmar2018tactile} while maintaining its low-cost and modular design.

\subsection{Dynamic Interactions For Long-term Engagement}

While enriching Lantern’s interactions through sound, language, and touch can deepen momentary engagement, sustaining this engagement over time requires more than additional modalities. Participants noted that static, pre-programmed behaviors limited Lantern’s sense of liveliness, underscoring the need for dynamic, adaptive interactions that evolve with use. Rather than relying on pre-scripted motion patterns, future iterations of Lantern should integrate real-time, adaptive behaviors driven by user inputs (\eg gestures, touch) or environmental sensing to create a more responsive, engaging and context-aware interaction experience \cite{altun2015recognizing}.

Designing behaviors for different phases of the interaction, particularly the start and end of interactions, could enhance sustained user engagement \cite{leite2013social}. For instance, Lantern could detect a user’s presence and initiate engagement with personalized cues, such as subtle movements, vibrations, or sounds that acknowledge the user and invite interaction. Similarly, the conclusion of an interaction should not feel abrupt but rather flow naturally, with behaviors like gradual motion deceleration or ambient sound cues (\eg a soft “exhale”) to provide closure. Additionally, adapting behaviors based on the interaction context—such as the time of day or ambient conditions—could make Lantern feel more responsive and alive.

By expanding Lantern’s behavior set \cite{antony2025xpress} and enabling real-time dynamic reactions, we can further explore how robotic objects can evoke deeper social engagement and interaction fidelity \cite{fink2014robot}.

\subsection{Limitations}

While Lantern and our exploratory work offer promising insights into the design and use of minimal robotic objects, certain limitations remain. As our studies primarily aimed to establish proof-of-concept, the findings reflect this exploratory focus, and further controlled research against baselines with inferential statistics is needed to isolate Lantern’s specific effects and better understand user needs and responses to robotic objects. Moreover, evaluating Lantern’s robustness, reliability, and maintainability in real-world settings through long-term deployments is a crucial next step. Future work should explore Lantern for the delivery of targeted longitudinal interventions in key domains such as sleep health \cite{antony2025social}, ADHD support \cite{o2024design}, mindfulness and meditation \cite{alimardani2020robot}.
  
Lantern primarily reduces economic barriers to HRI research; other obstacles persist. Not all researchers have access to tools like 3D printers or possess the technical expertise needed for circuit assembly or programming. Expanding accessibility through measures such as alternative build-kit (e.g., cardboard) and visual programming interfaces can broaden Lantern’s adoption and impact.




\begin{acks}
\textbf{Funding Source:} Johns Hopkins Malone Center for Engineering in Healthcare and National Science Foundation award \#2143704. 
\end{acks}

\bibliographystyle{ACM-Reference-Format} 
\bibliography{references}

\end{document}